\documentclass[lettersize,journal]{IEEEtran}

\usepackage{graphicx}
\usepackage{comment}
\usepackage{amsmath,amssymb} 
\usepackage{color}
\usepackage{booktabs}
\usepackage{epsfig}
\usepackage{graphicx}
\usepackage{amsmath}
\usepackage{amssymb}
\usepackage{amsmath, amssymb}
\usepackage{makecell}
\usepackage{multirow}
\usepackage{color}
\usepackage{graphicx}
\usepackage{amsmath}
\usepackage{amssymb}
\usepackage{booktabs}
\usepackage{color}
\usepackage{diagbox}
\usepackage{pifont}
\usepackage{multirow}
\usepackage{hyperref}
\begin{document}



\title{Generalizing to Out-of-Sample Degradations via Model Reprogramming}


\author{Runhua Jiang, Yahong Han
	\thanks{R. Jiang and Y. Han are with College of Intelligence and Computing, Tianjin University, Tianjin 300350, China (E-mail: ddghjikle@tju.edu.cn, yahong@tju.edu.cn). Corresponding author: Yahong Han.}
	\thanks{This work is supported by the NSFC (under Grant 62376186, 61932009). This work is also supported by the CAAI-Huawei MindSpore Open Fund.}
}

\maketitle
\begin{abstract}
	Existing image restoration models are typically designed for specific tasks and struggle to generalize to out-of-sample degradations not encountered during training. While zero-shot methods can address this limitation by fine-tuning model parameters on testing samples, their effectiveness relies on predefined natural priors and physical models of specific degradations. Nevertheless, determining out-of-sample degradations faced in real-world scenarios is always impractical. As a result, it is more desirable to train restoration models with inherent generalization ability. To this end, this work introduces the Out-of-Sample Restoration (OSR) task, which aims to develop restoration models capable of handling out-of-sample degradations. An intuitive solution involves pre-translating out-of-sample degradations to known degradations of restoration models. However, directly translating them in the image space could lead to complex image translation issues. To address this issue, we propose a model reprogramming framework, which translates out-of-sample degradations by quantum mechanic and wave functions. Specifically, input images are decoupled as wave functions of amplitude and phase terms. The translation of out-of-sample degradation is performed by adapting the phase term. Meanwhile, the image content is maintained and enhanced in the amplitude term. By taking these two terms as inputs, restoration models are able to handle out-of-sample degradations without fine-tuning. Through extensive experiments across multiple evaluation cases, we demonstrate the effectiveness and flexibility of our proposed framework. Our codes are available at \href{https://github.com/ddghjikle/Out-of-sample-restoration}{Github}.
\end{abstract}


\IEEEpeerreviewmaketitle
{}


\section{Introduction}
Image restoration has achieved remarkable success with the rapid development of deep neural networks. Previous researches \cite{liang2021swinir,zamir2022restormer,zhang2021single} always characterize specific types of degradations as individual issues and propose dedicated solutions. While this methodology effectively addresses some real-world scenarios, it falls short when considering more complex situations like autonomous driving on rainy days. In such cases, the perceived images can be degraded by a combination of rain, haze, blur, and noise, making it difficult to attribute the degradation to a specific form. Moreover, as real-world images can exhibit diverse and unpredictable degradation patterns, enumerating all possible degradations to train restoration networks is practically infeasible \cite{liu2022tape}. Therefore, it is imperative to develop restoration models that can effectively restore images degraded by various factors, including those not encountered during training.

\begin{figure}[t]
	\centering
	\includegraphics[width=\linewidth]{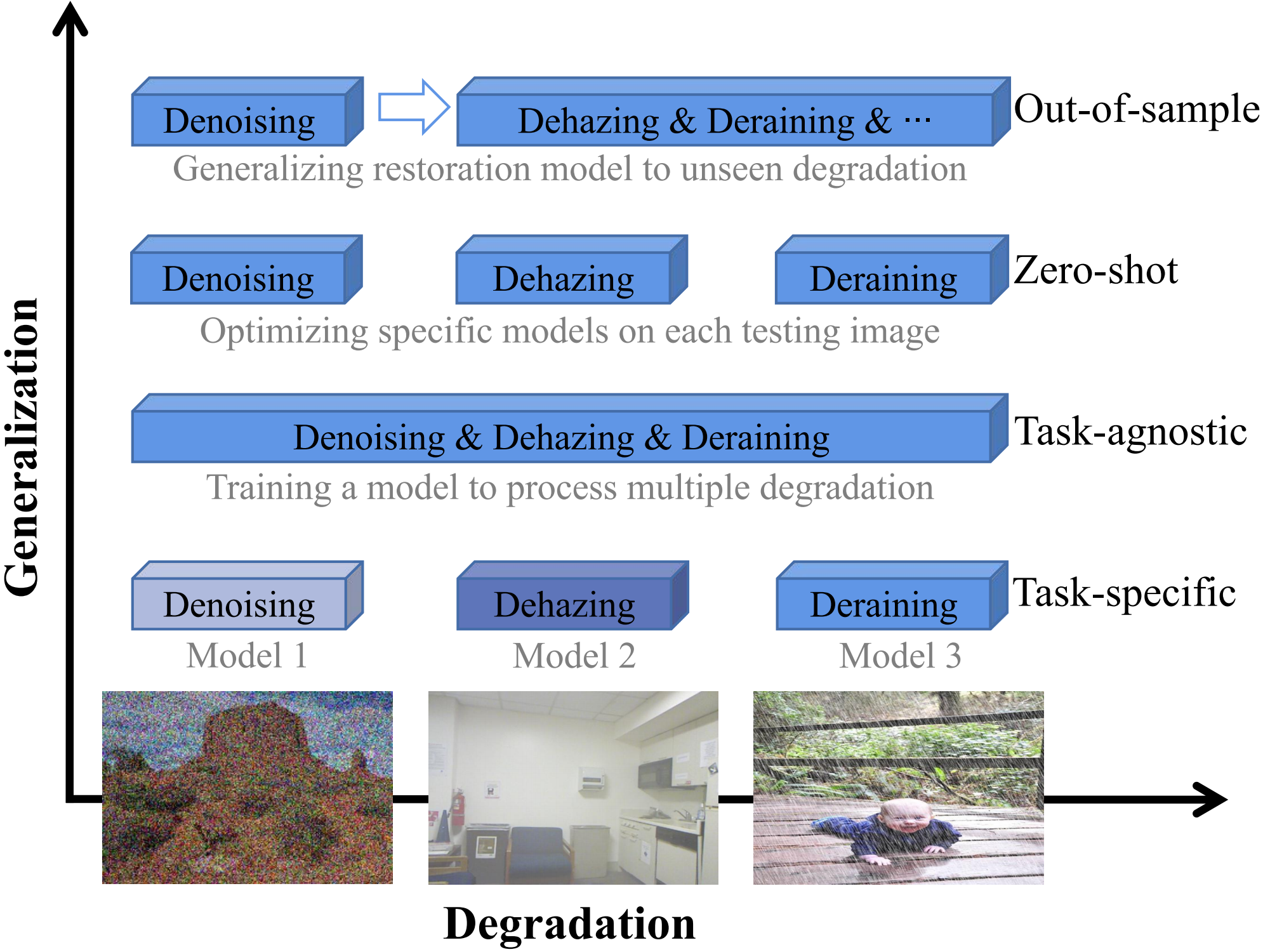}
	\caption{The introduced out-of-sample restoration task is to develop models with the capability of handling unknown degradations. It extends previous restoration researches as paying more attention to cross-degradation generalization.}
	\label{fig:motivation}
	\vspace{-0.2in}
\end{figure}

In contrast to supervised approaches focused on removing specific degradations, some studies explored the challenges of handling multiple degradations \cite{liang2021swinir,zamir2022restormer} or generalizing from synthetic images to real-world scenarios \cite{zhao2021refinednet}. To name a few, Liang \emph{et al.} \cite{liang2021swinir} introduced Swin Transformer layers to remove low-resolution, noisy, and compression artifacts. The authors of \cite{zamir2022restormer} proposed an efficient multi-head attention and feed-forward network for capturing long-range pixel interactions. Their experiments demonstrate that Transformer models, irrespective of degradation categories, can effectively perform when training and testing sets exhibit the same degradation. Conversely, other study \cite{zhao2021refinednet} highlights the domain gap between synthetic and real-world degraded images. They assume that both synthetic and real-world domains suffer from the same degradation, and perform unsupervised learning on the real-world domain. Recently, this motivation has been extended by zero-shot methods \cite{li2020zero,li2021you}, which assume the domain gap exists between training and testing sets rather than between synthetic and real-world training samples. To address this issue, zero-shot methods require prior knowledge of degradation categories, allowing them to fine-tune models for each testing sample. However, none of existing methods consider the challenging and practical circumstances wherein testing samples are affected by degradations not presented in the training samples.

\begin{figure}[t]
	\centering
	\includegraphics[width=\linewidth]{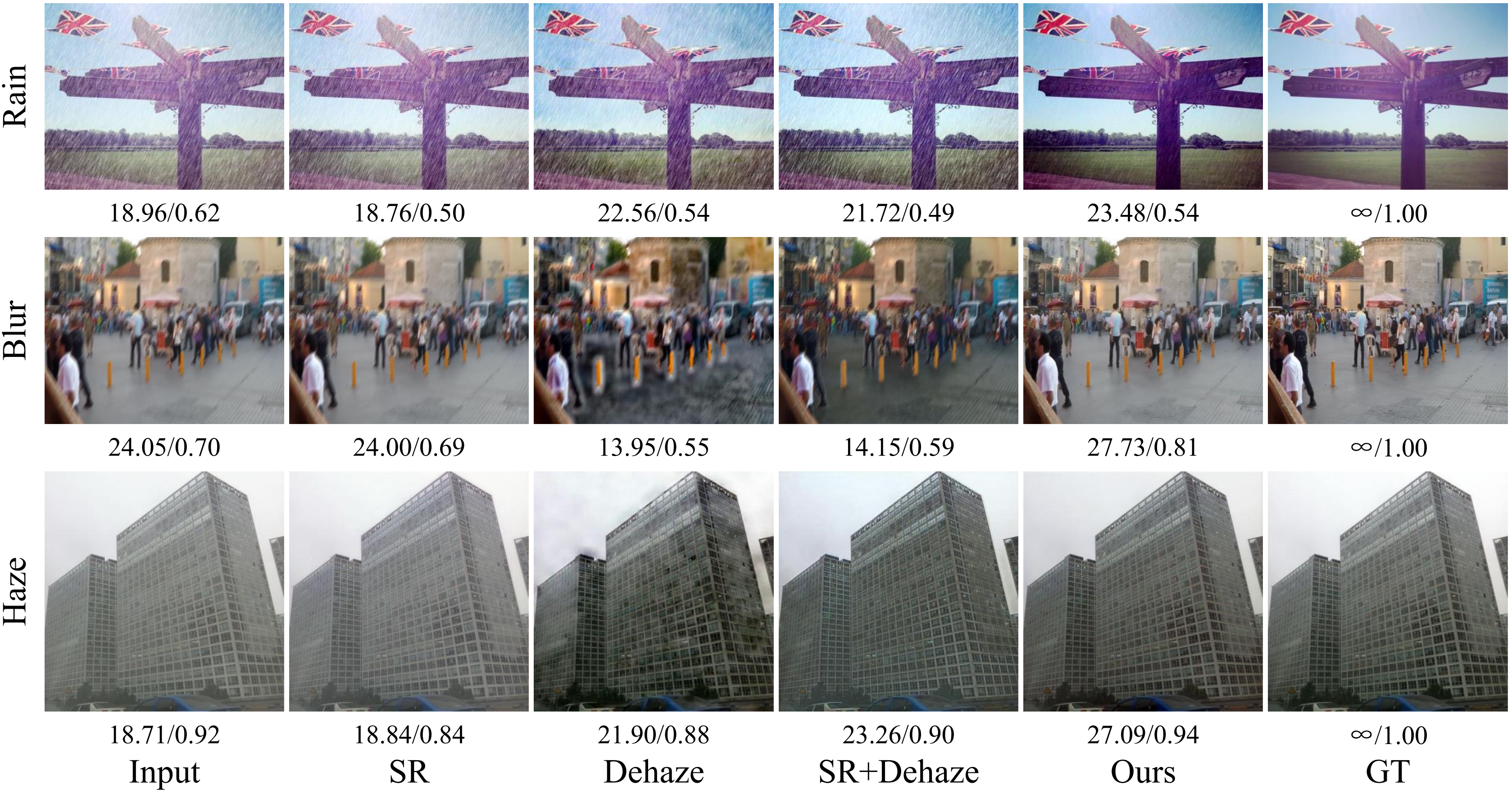}
	\caption{Visual examples of the out-of-sample degradation issue. These degraded images are restored by Super-Resolution (SR), Dehaze, SR+Dehaze networks, and our methods. PSNR and SSIM of each example are provided for better comparison.}
	\label{fig:visual_comp_self}
	\vspace{-0.2in}
\end{figure}

In this work, we introduce the Out-of-Sample Restoration (OSR) task, which aims to develop restoration models capable of generalizing to unknown degradations. Fig. \ref{fig:motivation} illustrates differences between this task and previous researches. Specifically, task-specific methods \cite{liang2021swinir,zhang2021single,li2017aod} focus on establishing image-to-image translation network for a certain kind of degradation, while task-agnostic methods \cite{liu2022tape} combine various types of degraded images to train a single restoration network. However, as Fig. \ref{fig:visual_comp_self} shows, while the task-specific network Dehaze effectively eliminates haze degradation, it struggles to address out-of-sample degradations such as blur and rain. Furthermore, the task-agnostic network SR+dehaze, trained on datasets encompassing super-resolution and dehazing networks, fails to exhibit improved performance on rainy examples. Although zero-shot researches \cite{li2020zero,li2021you} can address this issue by fine-tuning restoration models on testing samples, they require prior knowledge of degradation categories. In contrast, the OSR task aims to learn generalizable models from a limited set of training samples, making it different and complementary to previous researches.

While the OSR task shares similar motivation with domain adaptation task, which addresses the challenge of
multi-domain discrepancy, existing methods are ineffective in this context. The reason is that current domain adaptation methods are specifically designed for semantic-related tasks, such as image classification \cite{long2015learning,saito2018maximum} and semantic segmentation \cite{lengyel2021zero}. For example, Long \emph{et al.} \cite{long2015learning} embedded semantic representations of all task-specific layers into a kernel Hilbert space. Saito \emph{et al.} \cite{saito2018maximum} proposed to align the distributions of source and target domains by leveraging the semantic information within the classification boundaries. Subsequent studies, such as those by Li \emph{et al.} \cite{li2021semantic} and Han \emph{et al.} \cite{han2022learning},  divided feature space of classification networks according to their semantic discriminability or domain invariance. However, compared with semantic information, image content is more susceptible to degradation, resulting in complex discrepancies across different degradations.

An intuitive solution to the OSR task is to translate the out-of-sample degradations into degradations recognizable by restoration models, with the expectation that these models can then generate clear outputs. However, performing this translation in image space could raise complex image-to-image translation issues. To address this issue, we propose a novel approach by reprogramming restoration models using quantum mechanim. By this way, images are represented as wave functions comprising amplitude and phase terms. The amplitude encapsulates real-value features that represent the image content, while the phase modulates style vectors within a continuous space \cite{tang2022image,huang2022winnet}. These representations allow for efficient translation by aligning continuous phase vectors, addressing the complexities inherent in image-to-image translation. Consequently, our proposed framework is implemented with input and output transform functions. The input transform function utilizes phase-aware vision MLPs to disentangle input images into wave functions and align the phase intensity. Reprogrammed models can then efficiently translate the recognizable degradations into the style of clear images, and enhance content details within the amplitude component. Finally, the output transform function remaps these wave functions back into the image space. Through extensive experiments, it is demonstrated that our framework outperforms previous domain adaptation and image restoration methods in terms of generalization ability and restoration effectiveness.

Main contributions of this work can be summarized as follows:
\begin{itemize}
	\item We introduce a new task, \emph{i.e.}, Out-of-Sample Restoration (OSR), which requires models to handle a wide range of degradations, including those not encountered during training. This task complements previous restoration studies and focuses on developing models with cross-degradation generalization ability.
	
	\item To tackle the OSR task, we propose a model reprogramming framework that represents images as wave functions incorporating both amplitude and phase. This framework adapts restoration models to out-of-sample degradations by enhancing the amplitude element and transforming the phase term. To the best of our knowledge, this is the first attempt to combine model reprogramming with image restoration studies.
	
	\item Extensive experiments are conducted by using noisy, hazy, blurry, rainy, and low-resolution images. By comparisons with existing domain adaptation and image restoration methods, effectiveness and flexibility of the proposed framework is comprehensively verified.
\end{itemize}

\section{Related work}
In this section, we first provide an overview of image restoration. Then, the concepts of domain adaptation and model reprogramming are introduced as both of them are relevant to this work.

\subsection{Image Restoration}
Image restoration is to remove degradation artifacts from input images. Previous works \cite{qu2019enhanced,chen2021robust} characterize image restoration as specific tasks due to the complexity of real-world scenarios. For example, the image dehazing task aims at removing the haze degradation and enhance illumination appearances \cite{li2018benchmarking,zhang2021single}. The image deblurring task is to recover sharp image details from blurry artifacts. Learning-based models, such as convolution neural networks   \cite{liu2019griddehazenet} and Transformers \cite{liang2021swinir,zamir2022restormer}, have achieved remarkable success in addressing these specific tasks. Recent researches have also explored multi-task restoration, \emph{i.e.}, recovering multiple degradations through a task-agnostic network \cite{liu2022tape}. However, both task-specific and task-agnostic methods often struggle with degradations that are unseen during training, highlighting the need for more research on generalizable restoration models. Consequently, the zero-shot restoration task has been defined to transfer restoration models directly on testing images \cite{li2020zero,wang2022zero}. However, few attention has been paid to train generalizable restoration models with limited types of training images. In addition, zero-shot restoration does not consider influences caused by the transferred models and the target degradation. To address these issues, we introduce the Out-of-Sample Restoration (OSR) task, which provides a comprehensive understanding of the generalization ability of restoration models.
\begin{figure}[t]
	\centering
	\includegraphics[width=0.95\linewidth,height=0.8\linewidth]{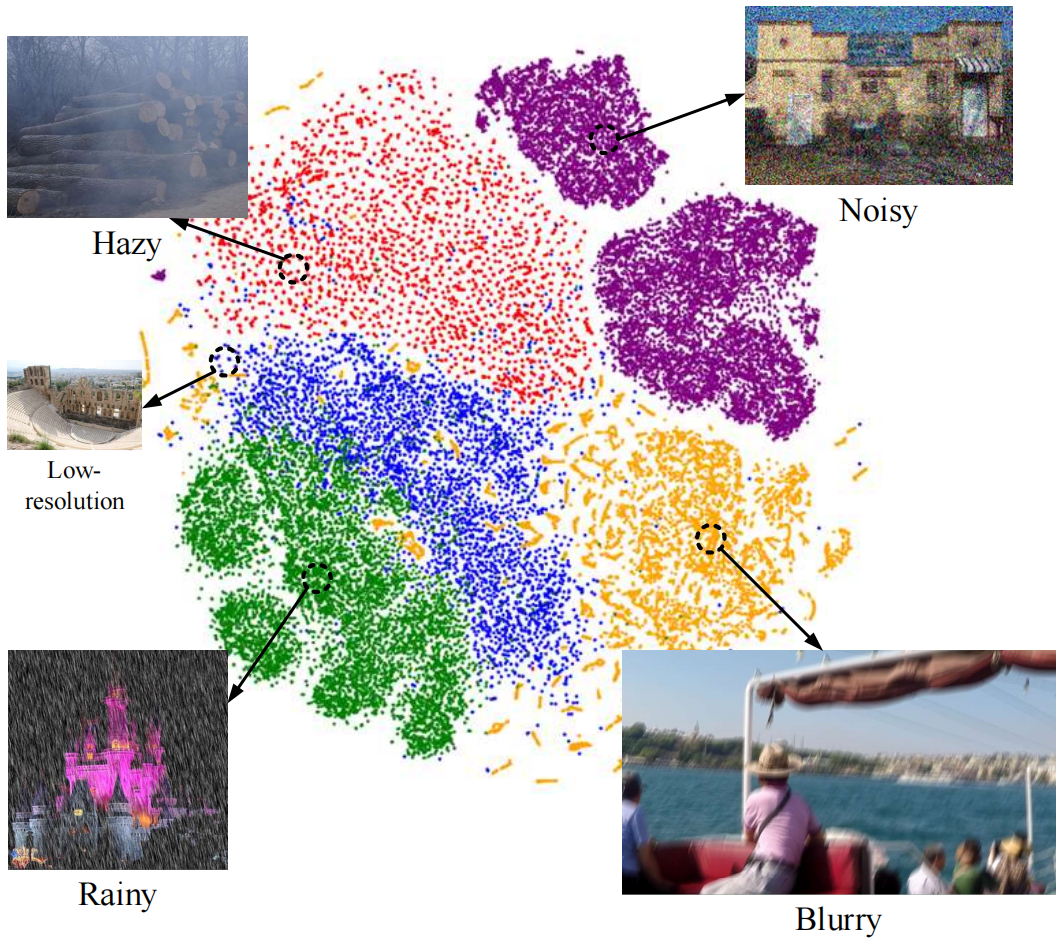}
	\caption{Distributions of images suffered from haze, noise, blur, rain, or low-resolution. These visualization results are obtained by leveraging t-SNE \cite{van2008visualizing} to visualize features generated by the perceptual function \cite{johnson2016perceptual}.}
	\label{fig:possibilioty}
	\vspace{-0.15in}
\end{figure}
\subsection{Domain Adaptation}
Domain adaptation aims at improving the performance of networks on target domains by transferring knowledge from different but related source domains. One popular methodology is to learn representations that are robust to domain discrepancy \cite{long2015learning,saito2018maximum}. To name a few, Chen \emph{et al.} \cite{chen2019transferability} introduced the concepts of feature transferability and discriminability through spectral analyses, demonstrating that eigenvectors with large singular values contribute to robustness against domain discrepancies. Latter, authors of \cite{li2021semantic} found that feature alignment should concentrate on semantic information for avoiding negative transfers. In contrast, Han \emph{et al.} \cite{han2022learning} considered the parameter space rather than the feature space. They demonstrated that different parts of network parameters are respectively related to transferability and discriminability. Thus, transferability-related parameters should be emphasized for learning domain-invariant features. However, existing domain adaptation methods are primarily designed for visual understanding tasks, \emph{e.g.}, image classification and semantic segmentation. As a result, they tend to perform poorly when it comes to learning transferable restoration models, motivating our exploration of the OSR task and its solutions.

\subsection{Model Reprogramming}
Model reprogramming refers to repurposing a pre-trained model from a source domain to solve tasks in a target domain without fine-tuning the model \cite{chen2022model}. It is also cast as adversarial reprogramming due to the implication that the task in the target domain can be different from pre-trained tasks \cite{elsayed2018adversarial}. To achieve this, an input transformation module is needed to map target-domain inputs to the input space of the pre-trained source model. Tsai \emph{et al.} \cite{tsai2020transfer} introduced the zero-padding operation with positional index for input transformation. During reprogramming, only the padded pixels can be changed. However, the performance of the repurposed model is influenced by the spatial size of the padded pixels. To alleviate this limitation, methods such as \cite{melnyk2022reprogramming} employ trainable encoders as the input transformation module. Then, cross-modal reprogramming is enabled by these trainable input transformation modules. For example, Neekhara \emph{et al.} \cite{neekhara2022cross} repurposed ImageNet classifiers for topic classification by using a set of trained tokens. Similarly, the Voice2Series proposed in \cite{yang2021voice2series} reprograms acoustic models for time series classification. Compared to the input transformation module, the output mapping layer is simpler due to that previous methods most consider classification tasks. Therefore, the output mapping can be achieved by manually defined strategies \cite{tsai2020transfer} or trainable linear layers \cite{neekhara2022cross}. In contrast to above methods, this work extents the model reprogramming to the image restoration community and defines the transform functions as wave-inspired disentanglement and fusion operations.


\begin{figure*}[t]
	\centering
	\includegraphics[width=0.95\linewidth]{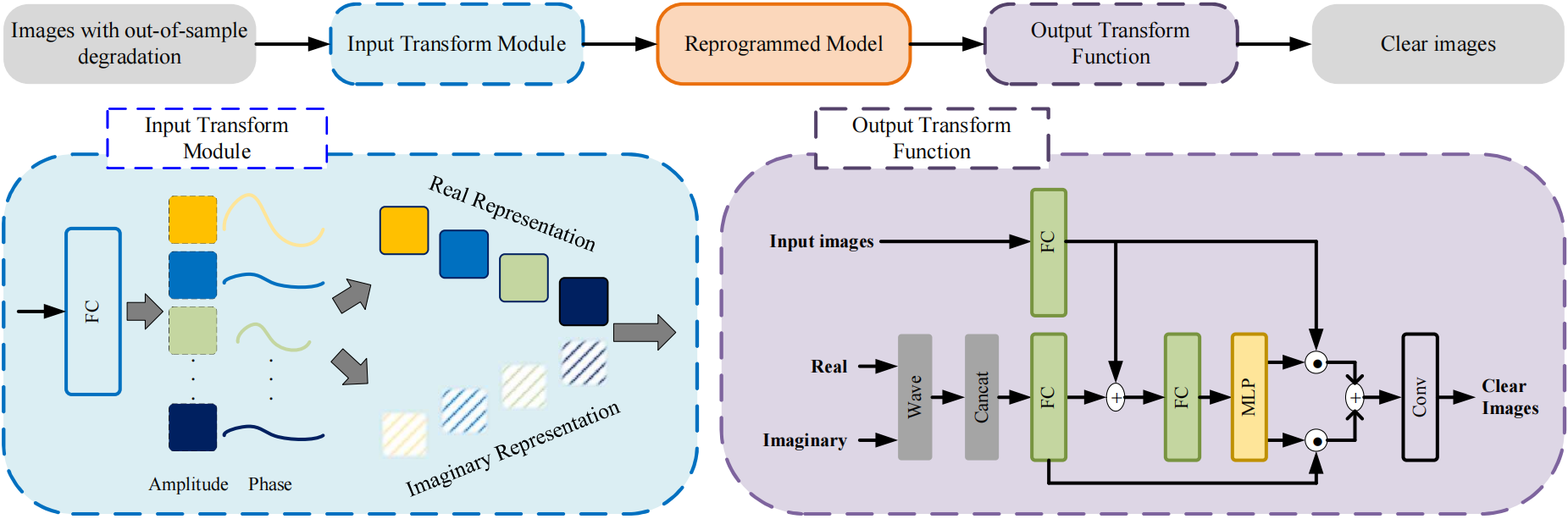}
	\caption{Overview of the proposed model reprogramming framework. The input transform module represents input images as amplitude and phase terms. The amplitude term is real-value feature representing image content, and the phase term modulates image styles. A restoration model is used to enhance the amplitude term and align the phase term. Finally, wave functions of processed amplitude and phase are formed and mapped to image space. Dashed rectangles in the top part represent trainable modules.
	}
	\label{fig:framework}
	\vspace{-0.2in}
\end{figure*}
\section{Proposed Approach}
\subsection{Task Formulation}\label{task_formulation}
As discussed previously, the out-of-sample restoration task aims to develop restoration models that can effectively handle degradations beyond training samples. To formally define this task, we consider five widely studied degradations: noise \cite{bychkovsky2011learning}, haze \cite{li2018benchmarking}, blur \cite{nah2017deep,shen2019human,su2017deep}, rain \cite{zhang2018density}, and low-resolution (LR) \cite{timofte2018ntire,agustsson2017ntire}. Then, the introduced OSR task can be defined as obtaining models capable of recovering all five degradations without training on each of them individually.

In general, there are two possible solutions for tackling this task. On the one hand, neural networks, equipped with abundant parameters, possess the possibility to learn substantial knowledge from available training samples. However, naively optimizing these over-parameterized networks easily over-fits training samples \cite{han2022learning}. In addition, due to distribution discrepancies shown in Fig. \ref{fig:possibilioty}, the augmentation of available degradations within training samples does not necessarily guarantee robust generalization to out-of-sample degradations. On the other hand, restoration models exhibit the capability to formulate mapping functions between specific types of degraded and clear images. Therefore, handling out-of-sample degradations becomes feasible after translating them into degradations that are recognizable by the restoration models. Nevertheless, the direct translation of degraded images introduces intricate image-to-image translation issues. Essentially, this method involves the reprogramming of restoration models through pre-processing operations, \emph{i.e.}, image-to-image translation. To navigate this challenge, we introduce the quantum mechanism and wave function, which are illustrated in the subsequent sections.

\subsection{Overall Framework}
To overcome the translation challenge of reprogramming restoration models, we propose to represent input images by quantum mechanics and wave functions. Fig. \ref{fig:framework} presents an overview of our proposed framework. First of all, the input transform module is designed following quantum mechanism, where entities are represented by wave functions comprising both amplitude and phase components \cite{tang2022image,arndt1999wave,heller1994scattering}. The amplitude corresponds to a real-valued feature that represents the maximum intensity of the wave, while the phase term modulates intensity patterns by indicating locations of each point. This design allows for the decoupling of input images into continuous vectors of content and style, with the style representation aligned to recognizable degradations of the reprogrammed model. Second, the reprogrammed model aims to map the style representation and enhance the content details by these components. Since no existing methods study the problem of reprogramming restoration models, two kinds of restoration models, \emph{i.e.}, randomly initialized and specifically trained, are explored in subsequent experiments. Finally, after processing these two components, the output transform function formulates wave functions and remaps them into the original image space to yield clear outputs.

\subsection{Input Transform Module}
According to quantum mechanism \cite{tang2022image,arndt1999wave,heller1994scattering}, an image can be represented as a wave $\hat{z}$ with both amplitude and phase terms: 
\begin{equation}
	\hat{z} = |z| \odot e^{\theta i},
	\label{eq:disentangle}
\end{equation}
where $|\cdot|$ and $\odot$ are absolute value operation and element-wise multiplication. $i$ denotes the imaginary unit satisfying $i^2 = -1$. The amplitude $|z|$ is a real-value feature representing image content, and the phase $\theta$ indicates distributions in the periodic function $e^{\theta i}$. In fact, the difference between amplitude  $|z|$ and real-value feature in neural networks is the absolute operation $|\cdot|$. Therefore, this operation can be integrated into the phase term as follows:
\begin{equation}
	|z| \odot e^{\theta i} = \left\{
	\begin{aligned}
		&z_t \odot e^{\theta_t i}, z_t >0,\\
		&z_t \odot e^{(\theta_t + \pi) i}, otherwise,\\
	\end{aligned}
	\right.
	\label{eq:absolute_operation}
\end{equation}
where $t$ denotes the position of each pixel. Therefore, the absolute operation can be removed and the amplitude can be obtained by a plain channel-FC operation with multi-channel inputs $[x_1, x_2, ..., x_n]$:
\begin{equation}
	\begin{aligned}
		z &= Channel{-FC}(x_i, W^c),\\
		&= W^c x_i, i=1, 2, ..., n,
	\end{aligned}
	\label{eq:amplitude}
\end{equation}
where $W^c$ is the weight with learnable parameters. Regarding the phase term $\theta$, which indicates the style distribution of the input images, various strategies can be employed to estimate it. One straightforward method is to use trainable vectors to represent it. However, this method does not consider diversity between images with different degradations. Therefore, following \cite{tang2022image}, we also adopt a channel-FC layer to estimate $\theta$.

Although above operations provide representations of the amplitude and phase terms, computations in the complex domain and the periodic function are challenging. Therefore, we further process these two terms using Euler's formulation and represent $\hat{z}$ with real part and imaginary part:
\begin{equation}
	\hat{z} = |z| \odot cos\theta + i |z| \odot sin\theta.
	\label{eq:mix}
\end{equation}
By doing so, a complex-value wave is represented as two real-value vectors, indicating the real and imaginary parts. As both of them are put through the reprogrammed model, these two vectors are defined with three channels and same spatial sizes as input images. Eq. \ref{eq:mix} not only alleviates the computational difficulties but also relaxes the requirement of accurately decomposing the phase and amplitude components, allowing the reprogrammed models to leverage their complementary relationships \cite{zhang2021single}.

\subsection{Output Transform Function}
With the input transform module decouples and adapts style representations of input images. The reprogrammed model is able to remove recognizable degradation and recover content details. Given outputs of the reprogrammed model, output transform function aims to remap wave-like representations into image space. To this end, the output transform function is designed as shown in Fig. \ref{fig:framework}. First, the real and imaginary terms restored by reprogrammed model are taken as inputs. To leverage their complementary relationships, these two terms are put through Eq. \ref{eq:mix}, resulting in wave representations that approximate clear images. By considering that the reprogrammed models tend to recover these two terms by different degrees, both of them and the wave representation are dynamically fused by a fully-connected layer. Owing to the matrix multiplication with learnable parameters, the wave representation can be further enhanced with more details and realistic colors. After that, the wave representation is dynamically added with input images to form residual information between degraded inputs and clear ground-truth. The added results are first processed by a fully-connected layer for channel-wise aggregation. Then, output features are spatially aggregated by MLP modules with two token-FC operations \cite{tang2022image}. Formally, the token-FC operation can be defined as
\begin{equation}
	\begin{aligned}
		o &= token{-FC}(x_i, W^t),\\
		&= W^t \odot x_i, i=1, 2, ..., n,
	\end{aligned}
	\label{eq:taken_FC}
\end{equation}
where $o$ indicates the output features. Above aggregation estimates pixel-wise consistencies between wave representation and input images. Therefore, results of the MLP modules are leveraged as dynamic weights to fuse wave representation and input images. Except the lightweight architecture, above functions are designed based on the quantum mechanism, leading to effective restoration for out-of-sample degradations.
\begin{figure}[t]
	\centering
	\includegraphics[width=\linewidth]{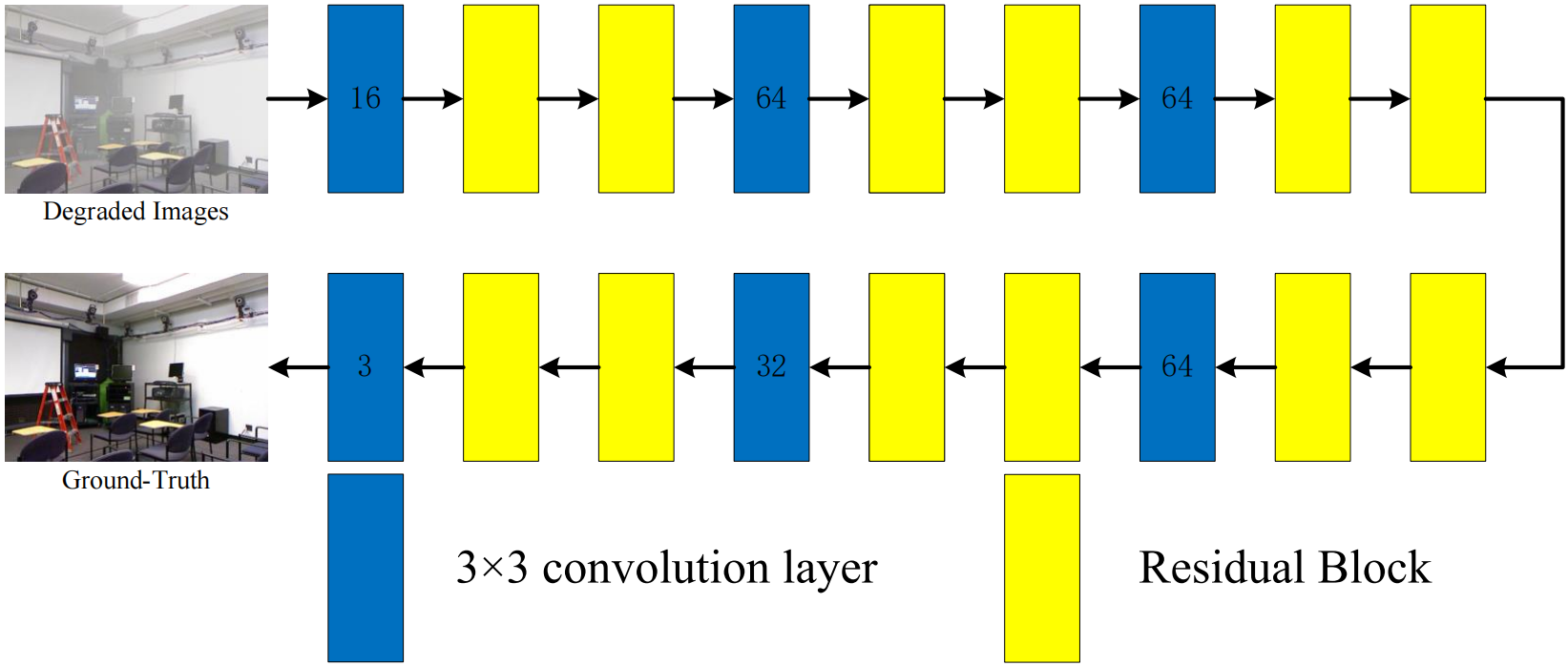}
	\caption{Architecture of the Res12 model. Numbers in blue rectangles denote channels of output features.}
	\label{fig:architecture}
	\vspace{-0.15in}
\end{figure}

For optimizing the input transform module and output transform function, the smooth $L_1$ loss and the perceptual loss are computed:
\begin{equation}
	\mathcal{L} = L_S + \lambda L_P,
	\label{eq:overall_objective}
\end{equation}
where $\lambda$ equals to 0.04 as \cite{liu2019griddehazenet} does. The smooth $L_1$ loss provides a quantitative measure of the difference between output images $\hat{x}$ and clear images $X$. It can be defined as:
\begin{equation}
	\begin{split}
		&L_S =\frac{1}{N}\sum_{x=1}^{N}\sum_{i=1}^{3}F_S(J_i(\hat{x})-J_i(X)),\\
		&F_S(x) = \left \{
		\begin{array}{ll}	
			0.5x^2,                    & if |x| < 1,\\
			|x| - 0.5,                                 & otherwise,
		\end{array}\right.
		\label{eq:l1}
	\end{split}
\end{equation}
where $J_i(\cdot)$ denote values of the $i$th channel of pixels in the output and clear images. $N$ is the total number of pixels. The perceptual loss \cite{johnson2016perceptual} quantifies the visual difference between output images and their clear versions:
\begin{equation}
	L_P =\sum_{j=1}^{3}\frac{1}{C_jH_jW_j}||\phi_j(\hat{x})-\phi_j(X)||_2^2,
	\label{eq:perceptual}
\end{equation}
where $\phi_j(\cdot)$ denotes three feature maps from VGG16 network \cite{johnson2016perceptual}. $C_j$, $H_j$, and $W_j$ specify the dimensions of $\phi_j(\cdot)$.


\begin{table}[t]
	\centering
	\fontsize{5.5}{8}\selectfont
	\tabcolsep=0.09cm
	\begin{tabular}{c|cccccc|cccccc}
		\toprule
		\multirow{2}{*}{\diagbox{Train}{Test}}&\multicolumn{6}{c|}{Res12}&\multicolumn{6}{c}{SwinIR \cite{liang2021swinir}}\\
		\cline{2-13}
		& LR & Rain & Noise & Blur & Haze & Avg & LR & Rain & Noise & Blur & Haze & Avg \\ \hline
		LR &\textbf{22.40}&\textcolor{blue}{20.67}&\textcolor{blue}{18.89}&25.35&12.12&19.89&\textbf{22.73}&\textcolor{blue}{19.82}&18.89&25.26&12.01&\textcolor{blue}{19.74}\\ 
		Rain &23.13&\textbf{32.21}&19.93&25.40&12.42&\textbf{22.62}&23.32&\textbf{32.29}&20.01&25.51&12.33&\textbf{22.69}\\ 
		Noise &23.60&22.63&\textbf{19.58}&25.43&12.36&20.72&23.52&22.38&\textbf{20.01}&25.47&\textcolor{blue}{11.99}&20.67\\ 
		Blur &23.19&22.15&19.92&\textbf{27.56}&\textcolor{blue}{12.06}&20.98&22.90&22.04&19.94&\textbf{27.23}&12.10&20.84\\ 
		Haze &\textcolor{blue}{15.94}&21.49&16.64&\textcolor{blue}{17.07}&\textbf{27.83}&\textcolor{blue}{19.79}&\textcolor{blue}{17.90}&22.46&\textcolor{blue}{17.68}&\textcolor{blue}{19.76}&\textbf{29.03}&21.37\\ 
		\bottomrule
	\end{tabular}
	\caption{PSNR metrics of Res12 and SwinIR \cite{liang2021swinir} trained on different kinds of degraded images. The best and the worst results are denoted by bold and blue fonts.}
	\label{table:baseline}
	\vspace{-0.25in}
\end{table}
\section{Experiment}
This section provides experimental analyses and comparison results. To be specific, we first illustrate datasets used for LR, rain, noise, blur, and haze degradations. Then, implementation details and evaluation protocols are introduced. After that, experiments about the OSR task and the proposed framework are discussed. Finally, the proposed framework is compared to existing domain adaptation and image restoration methods to validate its effectiveness.

\subsection{Dataset}
In this work, we form the out-of-sample restoration task by five widely researched degradations: low-resolution \cite{timofte2018ntire,agustsson2017ntire}, rain \cite{zhang2018density}, noise \cite{bychkovsky2011learning}, blur \cite{nah2017deep,shen2019human,su2017deep}, and haze \cite{li2018benchmarking}. 


\textbf{Low-resolution.} Following previous researches \cite{zamir2022restormer,timofte2018ntire}, we consider three upscaling factors, \emph{i.e.}, $\times 2$, $\times 3$, and $\times 4$. Training samples are obtained by down-sampling images from the DIV2K \cite{agustsson2017ntire} and Flickr2K \cite{timofte2018ntire} datasets, resulting in 10,650 training samples.  For evaluation, we use the B100 \cite{arbelaez2010contour} datasets, which have been processed with the same upscaling factors.

\textbf{Rain.} We utilize the Rain1200 \cite{zhang2018density} dataset, which contains images with three different levels of rain: heavy, medium, and light. This dataset includes 12,000 training images and 1,200 testing images. As these images have complex distributions across the different levels, we directly use these three kinds of images for both training and evaluation.

\textbf{Noise.} To generate noisy images, we employ MIT-Adobe Five5K \cite{bychkovsky2011learning} dataset following the experimental setup in \cite{gu2019self}. Specifically, we add additive Gaussian noise with standard deviations of 15, 25, and 50 to images in this dataset. As most compared methods do not take the standard deviation as input, noisy images with different standard deviations are not separated. For comparison, we obtain testing samples by adding Gaussian noise with same standard variation to CBSD68 \cite{roth2009fields} datasets.

\begin{table}[t]
	\centering
	\footnotesize
	\begin{tabular}{c|cccccc}
		\toprule
		Width & LR & Rain & Noise & Blur & Haze & Avg \\ 
		\hline
		128&\textcolor{blue}{22.95}&\textcolor{blue}{30.14}&\textcolor{blue}{19.68}&\textcolor{blue}{25.02}&12.44&\textcolor{blue}{22.05}\\
		64&\textbf{23.13}&\textbf{32.21}&\textbf{19.93}&25.40&\textcolor{blue}{12.42}&\textbf{22.62}\\
		32&23.05&31.54&19.85&25.39&\textbf{12.51}&22.47\\
		\midrule		
		Depth & LR & Rain & Noise & Blur & Haze & Avg \\ 
		\hline
		24&\textcolor{blue}{19.78}&32.03&\textcolor{blue}{16.96}&\textcolor{blue}{22.19}&\textbf{14.55}&\textcolor{blue}{21.10}\\
		12&\textbf{23.13}&\textbf{32.21}&\textbf{19.93}&\textbf{25.40}&\textcolor{blue}{12.42}&\textbf{22.62}\\
		6&23.09&\textcolor{blue}{31.26}&19.81&25.35&\textcolor{blue}{12.42}&22.39\\
		\midrule		
		Framework & LR & Rain & Noise & Blur & Haze & Avg \\ 
		\hline
		MindSpore \cite{huawei2022huawei}&\textbf{23.13}&\textbf{32.28}&\textbf{19.93}&\textbf{25.40}&\textbf{15.00}&\textbf{23.15}\\
		Pytorch&\textbf{23.13}&\textcolor{blue}{32.21}&\textbf{19.93}&\textbf{25.40}&\textcolor{blue}{12.42}&\textcolor{blue}{22.62}\\
		\bottomrule
	\end{tabular}
	\caption{Influences caused by architectures of Res12. For fair comparison, training samples are taken form the rain degradation. The best and the worst results are denoted by bold and blue fonts.}
	\label{table:architecture_res12}
	\vspace{-0.25in}
\end{table}
\textbf{Blur.} We utilize three popular datasets. The GoPro dataset \cite{nah2017deep} consists of 3,214 pairs of blurred images and ground-truth, captured from 33 sequences at a resolution of $1280 \times 720$. It is divided into 2,103 pairs for training and 1,111 pairs for testing. The VideoDeblurring dataset \cite{su2017deep} is a large-scale video deblurring dataset, where each video contains approximately 100 frames, and the frame size is $1280 \times 720$. The HIDE dataset \cite{shen2019human} includes two parts: HIDE I (1,304 long-shot pictures) and HIDE II (7,118 close-up pictures). We combine these three datasets to obtain 14,408 training samples. For quantitative evaluation, we compute metrics on the testing sets of the GoPro dataset.

\textbf{Haze.} The RESIDE dataset \cite{li2018benchmarking} contains both synthesized and real-world image pairs captured in indoor and outdoor scenarios.  It includes five subsets: Indoor Training Set (ITS), Outdoor Training Set (OTS), Synthetic Objective Testing Set (SOTS), Real World Task-driven Testing Set (RTTS) and Hybrid Subjective Testing Set (HSTS). We use 13,990 images from ITS as training samples. Additionally, we leverage 500 indoor and 500 outdoor images from the SOTS as testing samples.

\begin{table}[t]
	\centering
	\footnotesize
	\begin{tabular}{c|cccccc}
		\toprule
		\diagbox{Train}{Test} & LR & Rain & Noise & Blur & Haze & Avg \\ \hline
		LR \& Rain & 23.07&31.77&19.90&25.51&12.16&22.48 \\ 
		LR \& Noise &\textbf{23.33}&21.78&19.86&25.43&\textcolor{blue}{11.99}&20.48 \\ 
		LR \& Blur &23.01&22.12&19.81&27.24&12.04&20.84\\ 
		LR \& Haze &19.46&19.72&17.03&17.70&21.91&\textcolor{blue}{19.16}\\ 
		Rain \& Noise &23.12&\textbf{31.89}&19.91&25.42&12.46&22.56\\ 
		Rain \& Blur &23.10&31.59&\textbf{19.94}&\textbf{27.30}&12.16&22.82 \\ 
		Rain \& Haze &21.52&30.33&19.00&19.03&24.71&\textbf{22.92}\\ 
		Noise \& Blur &23.11&22.21&19.92&27.07&12.15&20.89 \\ 
		Noise \& Haze &\textcolor{blue}{18.19}&\textcolor{blue}{20.80}&\textcolor{blue}{16.59}&\textcolor{blue}{16.91}&\textbf{24.85}&19.47 \\ 
		Blur \& Haze &22.20&21.40&19.28&23.52&20.77&21.43\\ 
		\bottomrule
	\end{tabular}
	\caption{Baseline performance of Res12 model trained with various combinations of training images. The best and the worst results are denoted by bold and blue fonts.}
	\label{table:two_source_tasks}
	\vspace{-0.25in}
\end{table}
\begin{table}[t]
	\centering
	\tabcolsep=0.1cm
	\begin{tabular}{c|ccccccc}
		\toprule
		Translation&Res12&Perturbation \cite{neekhara2022cross}&Convolution&ResBlock\\
		\hline
		Avg. PSNR&19.16&14.86&12.16&7.67\\
		\bottomrule
	\end{tabular}
	\caption{Results of taking different translation modules.}
	\label{table:translation_space}
	\vspace{-0.2in}
\end{table}
\begin{figure}[t]
	\centering
	\includegraphics[width=\linewidth]{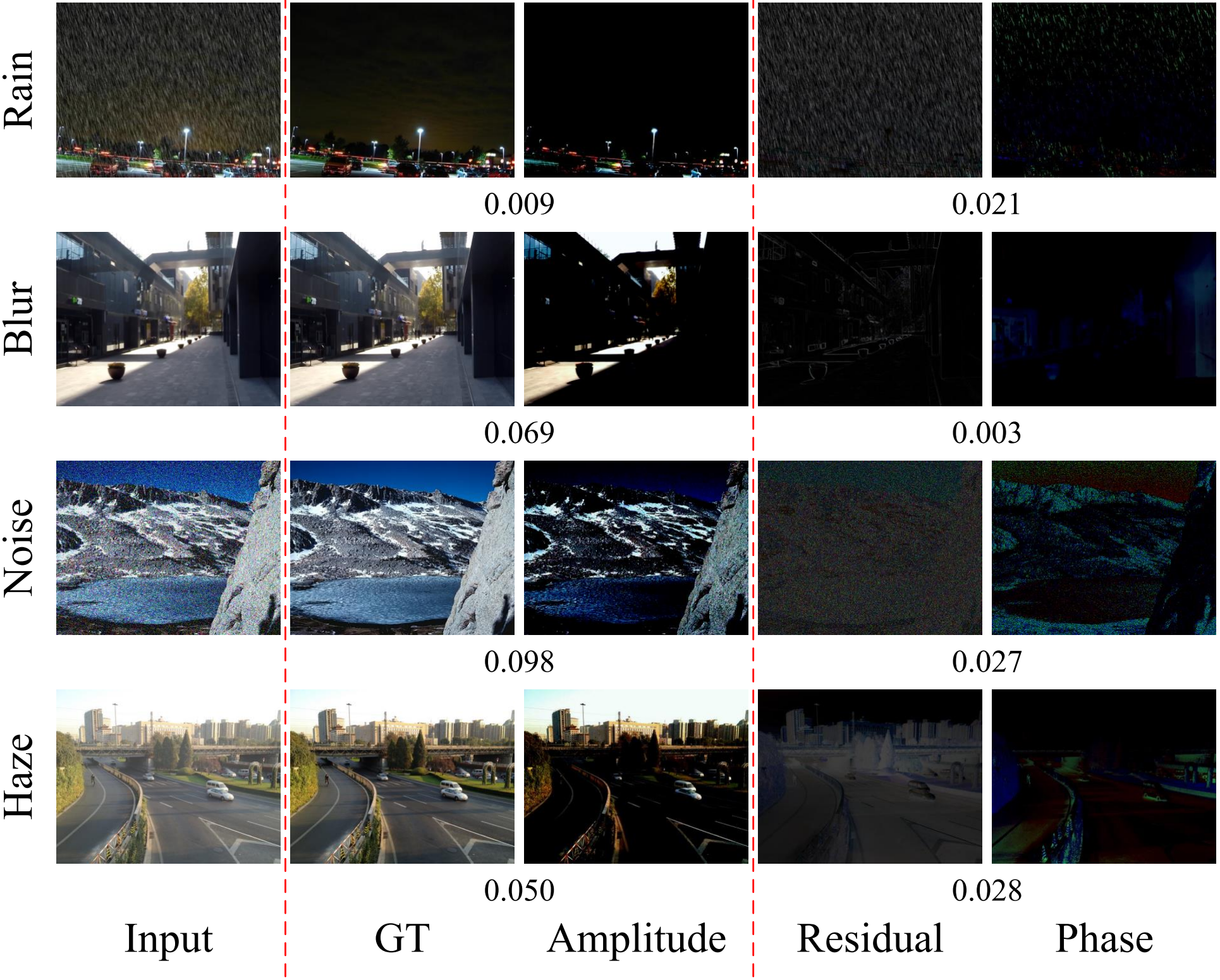}
	\caption{Visualizations of the amplitude and phase terms. The mean square error (MSE) is provided to demonstrate that amplitude and phase terms reflect image content and degraded styles, respectively.}
	\label{fig:model_analyses}
	\vspace{-0.2in}
\end{figure}
\begin{table}[t]
	\centering
	\begin{tabular}{c|ccccccc}
		\toprule
		Component&Amplitude&Phase&Amplitude\&Phase\\
		\hline
		Avg. PSNR&20.39&18.93&\textcolor{blue}{17.60}\\
		\hline
		Component&Imaginary&Real&Imaginary\&Real\\
		\hline
		Avg. PSNR&18.50&17.70&\textbf{20.84}\\
		\bottomrule
	\end{tabular}
	\caption{Results of leveraging different components of wave functions. The best and the worst results are denoted by bold and blue fonts.}
	\label{table:wave_functions}
	\vspace{-0.2in}
\end{table}
\subsection{Implementation}

\begin{table}[t]
	\centering
	\footnotesize
	\tabcolsep=0.1cm
	\begin{tabular}{c|cccccc}
		\toprule
		Methods&w/o Transform&ResBlock&WaveBlock \cite{tang2022image}&Channel-FC\\
		\hline
		Avg. PSNR&18.21&18.91&19.77&20.84\\		
		\bottomrule
	\end{tabular}
	\caption{Results of taking different input transform modules.}
	\label{table:input_transform_function}
	\vspace{-0.2in}
\end{table}
\begin{table}[t]
	\centering
	\footnotesize
	\tabcolsep=0.1cm
	\begin{tabular}{c|cc|ccc}
		\toprule
		Components&w/o Imaginary&w/o Real&MLP& MLP * 2& MLP * 4\\
		\hline
		Avg. PSNR&19.05&17.90&20.84&21.38&20.24\\
		\bottomrule
	\end{tabular}
	\caption{Results of taking different output transform functions. MLP * $n$ means $n$ MLP modules are implemented in this function.}
	\label{table:output_mix}
	\vspace{-0.25in}
\end{table}
\begin{figure}[t]
	\centering
	\includegraphics[width=0.97\linewidth]{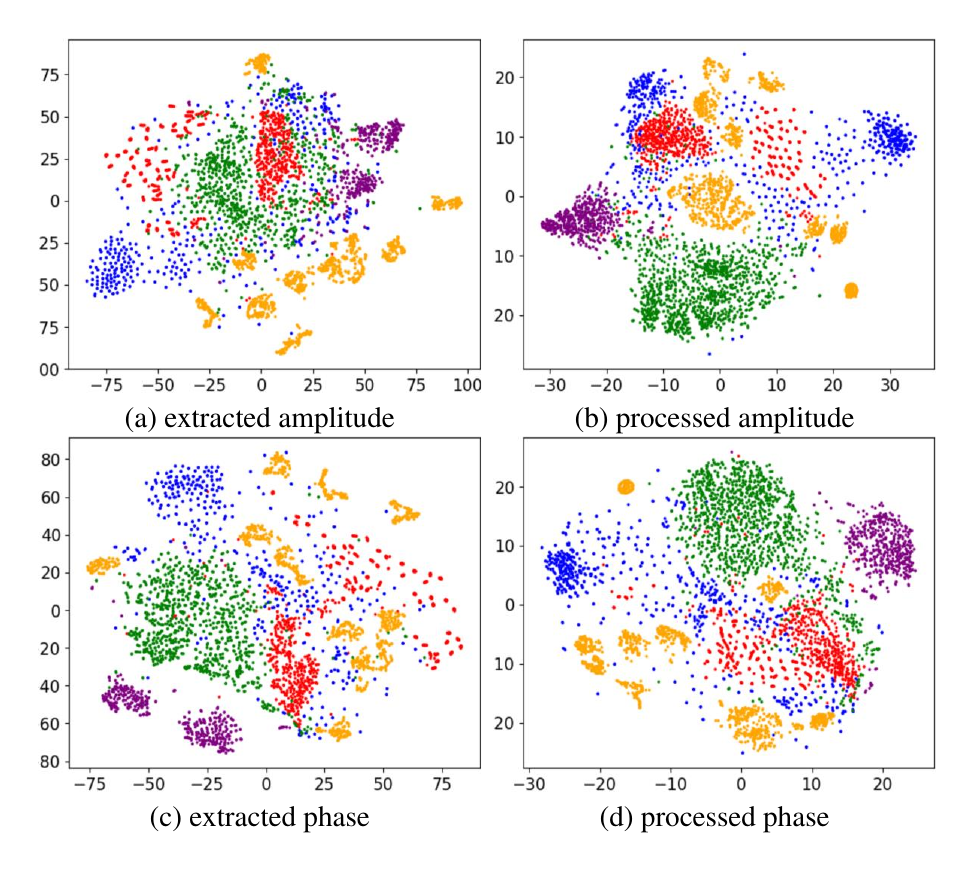}
	\caption{T-SNE \cite{van2008visualizing} results of amplitude and phase terms extracted from hazy, noisy, blurry, rainy, and low-resolution images. Components of these degraded images are represented by red, purpose, gold, green, and blue dots, respectively. Please note that ranges of axes in (b) and (d) are much smaller than that of (a) and (c).}
	\label{t_SNE_analyses}
	\vspace{-0.2in}
\end{figure}

\begin{table}[t]
	\centering
	\begin{tabular}{c|cc}
		\toprule
		Initialization Methods & Xavier-Normal \cite{glorot2010understanding} & Xavier-Uniform \cite{glorot2010understanding} \\
		\hline
		Avg. PSNR&18.70&17.54\\
		\hline
		Initialization Methods& Kaiming-Normal \cite{he2015delving} & Kaiming-Uniform \cite{he2015delving} \\
		\hline
		Avg. PSNR&17.33&20.84\\		
		\bottomrule
	\end{tabular}
	\caption{Results of reprogramming untrained models with different initialization methods.}
	\label{table:untrained}
	\vspace{-0.25in}
\end{table}

For the reprogrammed model, we adopt a plain convolution neural network (CNN) and term it as Res12. As Fig. \ref{fig:architecture} shows, it consists of six $3 \times 3$ convolution layers and 12 residual blocks. Each residual block comprises two convolution layers and a ReLU layer. These convolution layers map the input features to new dimensions such as 16, 64, or 32. By default, all convolution layers have a kernel size of $3 \times 3$, and the stride and padding sizes are set to 1. Transform functions in the proposed framework are trained using RGB image patches of size $120 \times 120$. For optimization, the Adam optimizer \cite{kingma2014adam} is used with a batch size of 8 for training samples. The initial learning rate is set to 0.001, and reduced by half every 20 epochs. All trainable parameters are optimized for 300 epochs, and all experiments are carried out on an NVIDIA GeForce GTX 2080Ti GPU via PyTorch and MindSpore \cite{huawei2022huawei} frameworks.

Due to degradation in the training set has important influences on model generalization, two evaluation protocols are leveraged in this work. First, the training set suffers from only one kind of degradation, and compared methods are evaluated on testing sets of all five degradations. Second, two kinds of degraded images are combined to obtain training samples, and all testing sets are leveraged to make comparison. To be more specific, according to baseline performance, 10,650 low-resolution images in DIV2K and Flickr2K, along with 13,990 images from the RESIDE dataset, are leveraged as training samples.

\subsection{Baseline Performance of OSR}
Since the OSR task is introduced in this work, our initial investigation focuses on assessing the performance of current CNN and Transformer models.

\textbf{Protocol 1.} First of all, we evaluate the baseline Res12 and popular Transformer model SwinIR \cite{liang2021swinir}, following the first evaluation protocol. This protocol employs only one kind of degraded images for training. As shown in Table \ref{table:baseline}, metrics calculated on out-of-sample degradation are significantly worse than metrics on the diagonal. These results demonstrate that current CNN and Transformer both suffer from discrepancies among various degradations. In addition, although SwinIR outperforms Res12 when training and testing sets share the same degradation, their average performances remain comparable. This suggests that the abundant parameters within SwinIR does not necessarily guarantee better generalization ability. Similar results can be observed from Table \ref{table:architecture_res12}, where Res12 with different widths and depths tend to have similar performance. We also explore to implement Res12 by the MindSpore \cite{huawei2022huawei} framework. Corresponding experiments provide comparable performance than the PyTorch framework, \emph{e.g.}, achieving 2.58 dB improvement on hazy samples. Overall, above experiments highlight two findings: 1) discrepancies across multiple degradations considerably influence evaluation results, and 2) over-parameterized restoration networks do not consistently exhibit superior generalization abilities.

\textbf{Protocol 2.} As discussed earlier, training restoration models with multiple types of degraded images seems to be a plausible solution for OSR. Therefore, we explore whether diverse training samples contribute to improve average performance. To this end, we combine two out of five degradations as training samples and evaluate the baseline model Res12. The results presented in Table \ref{table:two_source_tasks} indicate that above operation does not ensure performance improvement. Notably, combining LR and Haze as training samples results in an average PSNR decrease of approximately 4\%. However, the best performance is observed when combining images affected by rain and haze. Similar improvements are evident when rain or haze degradations are included in the training samples. These findings align with the earlier discussions about the OSR task, and emphasize the crucial effect of diverse discrepancies among different degradations.

\subsection{Reprogramming Framework}
With the second evaluation protocol and above baseline performances, here we validate effectiveness of the proposed framework.

\begin{figure}[t]
	\centering
	\includegraphics[width=\linewidth]{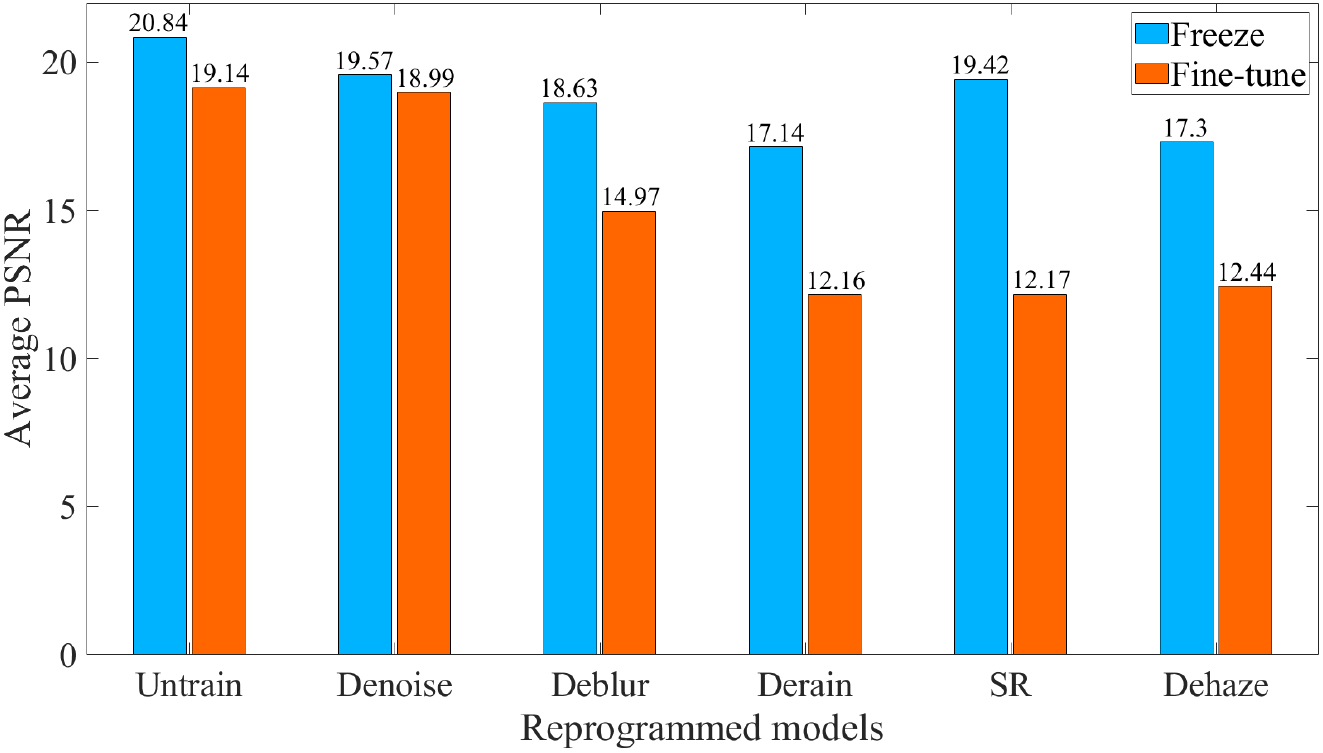}
	\caption{Reprogramming untrained or trained Res12 by fine-tuning or freezing model parameters.}
	\label{fig:victim_models}
	\vspace{-0.2in}
\end{figure}
\begin{table}[t]
	\centering
	\footnotesize
	\begin{tabular}{c|cccccc}
		\toprule
		Method & LR & Rain & Noise & Blur & Haze & Avg\\
		\hline
		Res12&19.46&19.72&17.03&17.70&21.91&19.16\\
		DAN \cite{long2015learning}&18.19&22.54&16.69&17.48&\textbf{27.94}&20.57\\
		DeepCoral \cite{sun2016deep}&20.36&20.33&17.75&18.59&20.09&19.42\\ 
		MCD \cite{saito2018maximum}&20.23&20.31&18.24&19.51&20.66&19.79\\
		BSP \cite{chen2019transferability}&20.34&20.27&18.30&18.72&20.60&19.65\\
		BNM \cite{cui2020towards}&18.93&\textbf{22.85}&16.52&17.78&27.12&20.64\\
		CIConv \cite{lengyel2021zero}&\textcolor{blue}{13.11}&\textcolor{blue}{11.64}&\textcolor{blue}{11.83}&\textcolor{blue}{12.47}&\textcolor{blue}{13.32}&\textcolor{blue}{12.47}\\
		SCDA \cite{li2021semantic}&20.12&19.18&17.23&18.87&21.59&19.40\\
		LTH \cite{han2022learning}&18.49&22.52&15.95&17.81&26.32&20.22\\
		\hline
		Ours&\textbf{22.04}&22.47&\textbf{19.41}&\textbf{22.51}&17.75&\textbf{20.84}\\
		\bottomrule
	\end{tabular}
	\caption{Performance comparisons with previous domain adaptation methods. The best and the worst results are denoted by bold and blue fonts.}
	\label{table:comp_domain_adaptation}
	\vspace{-0.3in}
\end{table}
\textbf{Translation Space.} As discussed in Sec \ref{task_formulation}, although translating out-of-sample degradation into recognizable categories can empower restoration models handle out-of-sample degradation, conducting such translation in the image space inevitably introduces additional challenges. To answer this issue, our reprogramming framework decouples and translates out-of-sample degradation by quantum mechanism. Here we validate above discussion by experiments in Table \ref{table:translation_space}. Firstly, following previous reprogramming research \cite{neekhara2022cross}, we introduce an additional noise perturbation to translate input images with out-of-sample degradation. Compared to the baseline Res12, this translation method results in an average PSNR decrease of 4.3 dB. Moreover, cascade convolutions and residual blocks are implemented to translate out-of-sample degradations in the image space, denoted as the Convolution and ResBlock methods. Notably, the image-to-image translation of out-of-sample degradation significantly decreases overall performance, aligning with our earlier discussion. 

On the other hand, different components of the wave function are put through the reprogrammed model. As Table \ref{table:wave_functions} shows, it is evident that solely utilizing imaginary or real representations yields inferior performance compared to the Imaginary\&Real method. Moreover, the direct utilization of amplitude or phase components results in a degradation in performance compared to the Imaginary\&Real approach, owing to the intricate computations involved in the complex domain and periodic functions. This trend persists when taking only amplitude or phase as inputs. Furthermore, a noteworthy observation is the superior performance of amplitude over the phase term. This underscores the fact that the amplitude contains abundant content information of the ground-truth, whereas the phase term primarily captures image styles. Overall, above experimental analyses demonstrate the pivotal role of amplitude in encoding the majority of the ground-truth content, while the phase term is more focused on encoding image styles.

\textbf{Input Transform Module.} After analyzing the imaginary and real representations, we proceed to verify the efficacy of our input transform module. To this end, we conduct experiments by removing this module or replacing it with residual blocks and wave blocks \cite{tang2022image}. As Table \ref{table:input_transform_function} indicates, the w/o Transform method, obtained by removing this module, results in worse performance than baseline Res12. The reason is that 1) the output transform function cannot perform effectively without real and imaginary representations provided by this module, and 2) the reprogrammed model suffers from the cross-degradation discrepancies. Moreover, it is observed that both residual block and wave block fail to improve the average PSNR. Although both wave block and our module aim to obtain real and imaginary representations, the wave block further mixes these two representations as output features, presenting challenges in recovering image content and style from corresponding representations.

We further validate the input transform module through visualization results. Since the input images are generated by degrading the image content within the ground-truth (GT), we utilize the GT and residual images, obtained by comparing the input and GT images, as benchmarks for assessing image content and degraded appearances. For clear comparison, we further measure the mean square error (MSE) between GT and amplitude to verify their content consistency. On the other hand, since the residual image reflects influence caused by degradation factors, we evaluate the MSE between this residual image and the phase term to ascertain whether the phase term effectively captures the styles of different degradations. As Fig. \ref{fig:model_analyses} shows, the extracted amplitude and phase terms exhibit subtle distinctions and maintain a similar appearance to the GT and residual images. Therefore, it is verified that amplitude encodes the image content and phase encodes the image style. To demonstrate that our methods can adapt the phase term and enhance the amplitude term, we provide Fig. \ref{t_SNE_analyses} to illustrate distributions of these two terms. By examining the ranges of axes in these sub-figures, it can be found that our methods generate phase and amplitude terms with significantly tighter distributions. This observation underscores the effectiveness of our approach in recovering out-of-sample degradations.

\textbf{Output Transform Function.} Next, we validate effectiveness of the output transform function. As can be observed from Table \ref{table:output_mix}, removing the imaginary or real components both degrades effectiveness of the output transform function. In addition, results in the right part indicate that the average performance can be improved by employing more MLP layers. By repeating two MLP modules \cite{tang2022image}, the average performance increases to 21.38. However, when numbers of the MLP modules become 4, the average performance becomes worse than employing single MLP.

\subsection{Reprogrammed Models}
As our framework aims to reprogram restoration models to out-of-sample degradations, here we investigate influences of different reprogrammed models. To this end, two kinds of Res12 are leveraged as the reprogrammed model, \emph{i.e.}, randomly initialized models and trained on one kind of degraded images (presented in Table \ref{table:baseline}).

\textbf{Fine-tuned \emph{v.s.} Frozen.} Fine-tuning pre-trained models on downstream tasks is a popular methodology in transfer learning. Therefore, we first validate whether fine-tuning is beneficial for reprogramming restoration models. From Fig. \ref{fig:victim_models}, it can be observed that for reprogramming restoration models, freezing model parameters proves to be more effective than fine-tuning them. This effectiveness can be attributed to three reasons. First, model reprogramming and fine-tuning are two distinct tasks. Second, during reprogramming, optimizing model parameters increases the difficulty in learning effective transformation modules. Third, discrepancies among out-of-sample degradations are challenging for fine-tuning. According to the experimental results, the restoration models are frozen during reprogramming.

\textbf{Trained \emph{v.s.} Untrained.} Following previous model reprogramming researches \cite{neekhara2022cross,chen2022model}, here we explore differences between reprogramming trained and untrained restoration models. As depicted in Fig. \ref{fig:victim_models}, reprogramming untrained models yields superior performance. Although this seems incredible, similar results can also be found in \cite{neekhara2022cross,chen2022model,li2021you}. The underlying reason lies in the mechanism of model reprogramming. Untrained models possess unorganized knowledge that is not specific to any particular task, making them less robust against adversarial noises. Moreover, we consider four different initialization methods. Xavier initialization \cite{glorot2010understanding} is derived from the assumption that the singular values of the Jacobian associated with each layer should close 1. He \emph{et al.} \cite{he2015delving} find that this assumption is not suitable for non-linear layers. They thus proposed a robust initialization method that particularly considers the rectifier nonlinearities. Table \ref{table:untrained} shows untrained models with different initialization methods also have diverge performance.

\begin{figure}[t]
	\centering
	\includegraphics[width=\linewidth]{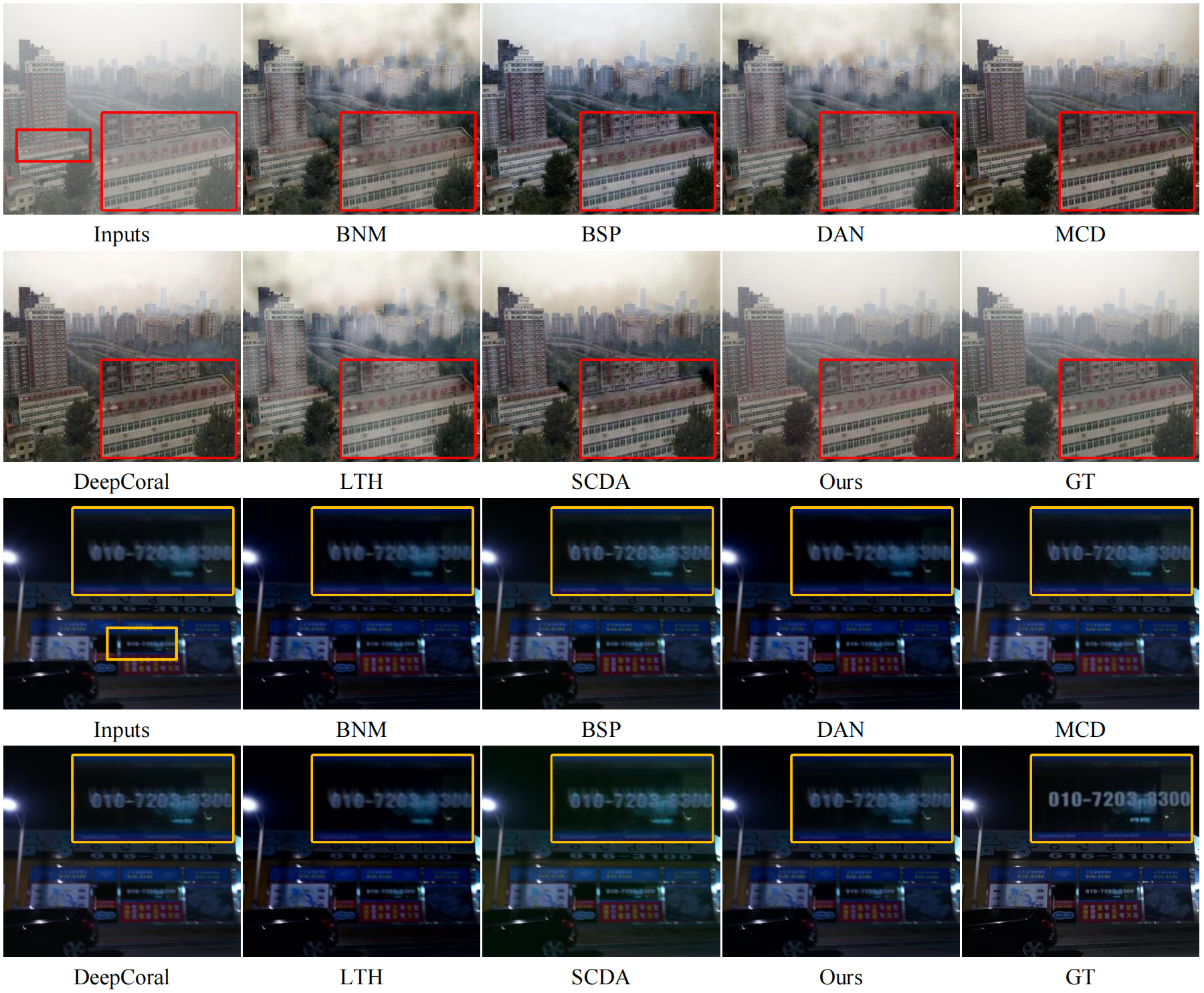}
	\caption{Visual comparisons between the proposed framework and previous domain adaptation methods \cite{cui2020towards,chen2019transferability,long2015learning,sun2016deep,han2022learning,li2021semantic} on image dehazing and deblurring tasks.}
	\label{fig:visual_comp_DA}
	\vspace{-0.25in}
\end{figure}
\begin{figure*}[t]
	\centering
	\includegraphics[width=\linewidth,height=0.41\textheight]{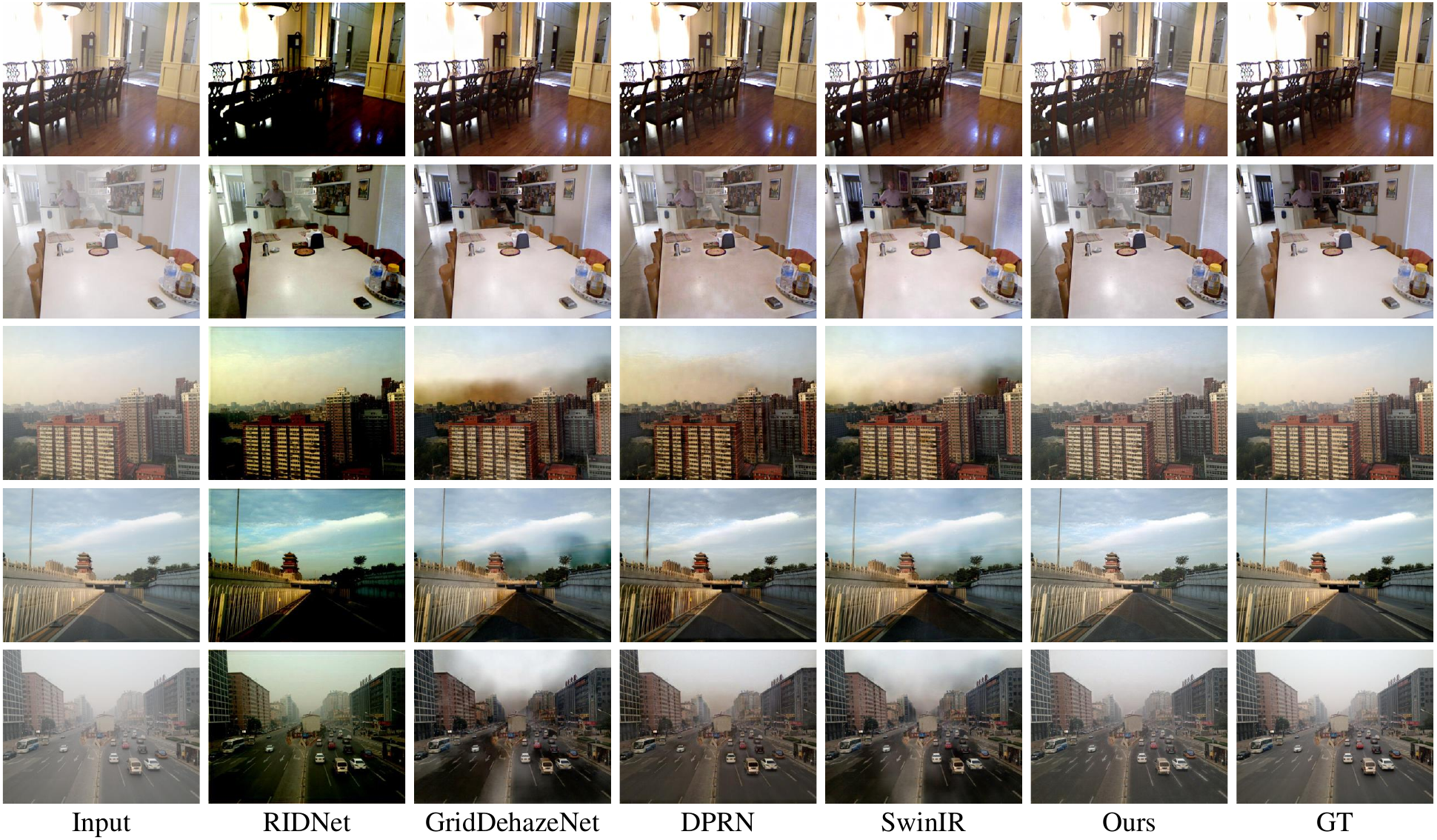}
	\caption{Visual comparisons on the image dehazing task \cite{anwar2019real,liu2019griddehazenet,zhang2021single,liang2021swinir}. The top two examples are indoor hazy images, while other examples are outdoor hazy images. It is worth noting that these outdoor hazy scenarios do not exist in training sets.}
	\label{fig:visual_comp_dehaze}
	\vspace{-0.15in}
\end{figure*}
\begin{table}[t]
	\centering
	\fontsize{6}{10}\selectfont
	\tabcolsep=0.1cm
	\begin{tabular}{c|c|cccccc|c}
		\toprule
		Taxonomy&Method & LR & Rain & Noise & Blur & Haze & Avg &Param\\
		\hline
		\multirow{3}{*}{CNN}&Res12&19.46&19.72&17.03&17.70&21.91&19.16&\textbf{7.39}\\
		&MPRNet \cite{zamir2021multi}&16.36&19.17&14.58&16.46&21.91&17.70&157.41\\
		&MIRNet \cite{zamir2020learning}&15.63&22.87&16.67&18.30&28.68&20.43&317.88\\
		\hline
		\multirow{2}{*}{Transformer}&SwinIR \cite{liang2021swinir}&16.63&22.69&17.32&18.52&29.56&20.94&19.05\\
		&Restormer \cite{zamir2022restormer}&17.64&22.26&15.65&21.66&16.90&18.82&117.40\\
		\hline
		\multirow{2}{*}{Attention}&GridDehazeNet \cite{liu2019griddehazenet}&15.45&22.26&16.45&20.94&30.73&21.17&9.58\\
		&RIDNet \cite{anwar2019real}&\textcolor{blue}{5.99}&\textcolor{blue}{8.59}&\textcolor{blue}{5.79}&\textcolor{blue}{6.43}&12.17&\textcolor{blue}{7.79}&15.00\\
		\hline
		\multirow{3}{*}{Formulation}&DPRN \cite{zhang2021single}&17.60&19.55&16.07&17.62&19.45&18.06&85.44\\
		&DerainRLNet \cite{chen2021robust}&18.33&21.81&15.57&19.44&21.81&19.39&58.22\\
		&Syn2Real \cite{yasarla2020syn2real}&14.50&22.80&13.97&15.54&22.07&17.78&26.18\\
		\hline
		\multirow{3}{*}{Unsupervised}&CVF-SID \cite{neshatavar2022cvf}&15.85&21.22&\textbf{24.16}&\textbf{25.37}&\textcolor{blue}{11.96}&19.71&11.86\\
		&D4 \cite{yang2022self}&7.85&12.27&9.61&9.66&15.86&11.05&\textcolor{blue}{284.29}\\
		&YOLY \cite{li2021you}&13.66&19.17&15.00&16.20&19.38&16.68&236.58\\
		\hline
		Diffusion&WeatherDiffusion \cite{ozdenizci2023restoring}&13.01&11.92&11.96&12.21&24.04&14.63&109.74\\		
		\hline
		\multirow{5}{*}{Reprogramming}&Res12&\textbf{22.04}&22.47&19.41&22.51&17.75&20.84&\textbf{7.39}\\
		&GridDehazeNet&17.32&23.01&16.93&18.24&\textbf{31.05}&21.31&9.58\\
		&SwinIR&16.52&22.74&17.35&18.50&29.54&20.93&19.05\\
		&MPRNet&17.80&20.18&16.20&16.59&20.10&18.17&157.41\\
		&MIRNet&18.63&\textbf{23.56}&18.08&19.25&27.39&\textbf{21.38}&317.88\\
		\bottomrule
	\end{tabular}
	\caption{Performance comparisons with previous methods following the Protocol 2, \emph{i.e.}, training on hazy and LR images but evaluating on five kinds of degraded images. The best and the worst results are denoted by bold and blue fonts. Param indicates overall parameters ($\times 10^5$) of each model.}
	\label{table:comp_image_restoration}
	\vspace{-0.35in}
\end{table}
\subsection{Comparisons}
\textbf{Domain Adaptation Methods.} For evaluating generalization ability of the proposed framework, we first compare it with domain adaptation methods. Most of these methods are proposed as specific loss functions, which need access to both source and target samples. Therefore, these functions together with Eq. \ref{eq:overall_objective} are used to train the baseline Res12. Table \ref{table:comp_domain_adaptation} presents corresponding results. It is evident that our framework is more suitable for restoration models. Compared with the latest method LTH \cite{han2022learning}, the proposed framework achieves better performance on out-of-sample degradations such as noise and blur. Fig. \ref{fig:visual_comp_DA} further illustrates that previous domain adaptation methods are ineffective to learn transferable restoration models. As a result, these compared methods cannot recover haze degradation and out-of-sample blur artifacts.

\textbf{Image Restoration Methods.} The proposed framework is further compared with image restoration methods. Table \ref{table:comp_image_restoration} presents the comparison with Transformer-based \cite{liang2021swinir, zamir2022restormer} and Diffusion-based \cite{ozdenizci2023restoring} methods, all trained on haze and LR degradations but tested on five restoration circumstances. By reprogramming Res12, although our method may not surpass over-parameterized methods like SwinIR \cite{liang2021swinir} and GridDehazeNet \cite{liu2019griddehazenet}, it significantly improves the generalization ability of the Res12. In addition, when the reprogrammed model is replaced by strong baselines such as GridDehazeNet, our framework achieves better performance. It can be observed that our methods can effectively enhance the generalization ability of MPRNet \cite{zamir2021multi} and MMIRNet \cite{zamir2020learning}, all without significantly causing additional parameters. Specifically, upon reprogramming, the PSNR metrics of MPRNet witness an increase of approximately 7.8\% across LR, rain, noise, and blur images. Similarly, leveraging our reprogramming framework results in 9.0\% improvement for the strong baseline MIRNet across the same set of images. These results and Fig. \ref{fig:visual_comp_dehaze} underscore the efficacy of our methods in augmenting the performance of various restoration models. 


In addition to performance improvements, Table \ref{table:comp_image_restoration} also highlights two important observations. First, among the 14 compared methods, only Restormer \cite{zamir2022restormer}, DPRN \cite{zhang2021single}, DerainRLNet \cite{chen2021robust}, Syn2Real \cite{yasarla2020syn2real}, and CVF-SID \cite{neshatavar2022cvf} exhibit inferior performance on the haze degradation. DerainRLNet and Syn2Real are specialized for image deraining, while CVF-SID is designed for image denoising, thereby resulting in poorer performance in haze removal compared to their respective specialized tasks. In contrast, DPRN's design, based on physical models of rainy and hazy images, yields similar performance levels for rain and haze degradations. The spatially enriched depth-wise convolution layers make Restormer more sensitive to depth-wise and spatially-variant degradations, \emph{i.e.}, rain and blur. These results indicate that model architectures are related to the generalization ability. Second, upon reprogramming, the Res12, MPRNet, and MIRNet tend to suffer from sacrificed performance in dehazing. Since the parameters of these baselines remain unchanged, applying the baselines directly can yield better performance on haze and LR degradations. Such results underscore the importance of the trade-off between generalization and existing performance. Consequently, it is imperative to investigate how the parameter space of restoration models influences performance on both within-training-set and out-of-sample degradations.

\section{Conclusion}
This work explores the challenging task of out-of-sample restoration, aiming to develop restoration models capable of generalizing to degradations not encountered during training. To tackle this task, we propose a model reprogramming framework inspired by image-to-image translation and quantum mechanics. In this framework, images are transformed into wave functions represented by amplitude and phase terms, which capture degradation-invariant image content and degradation-specific style distributions, respectively. The process of recovering out-of-sample restoration then involves enhancing the amplitude term and aligning the phase component. Instead of training restoration networks from scratch, our framework leverages existing restoration models and adapts them using two transform functions. Extensive experimental results demonstrate effectiveness and flexibility of the proposed framework.

{\small
	\bibliographystyle{IEEEtranTIE}
	\bibliography{myreferences}
}
\end{document}